\documentclass[10pt,twocolumn,letterpaper]{article}

\usepackage{cvpr}              

\usepackage{graphicx}
\usepackage{amsmath}
\usepackage{amssymb}
\usepackage{booktabs}
\usepackage{blindtext}
\usepackage{caption}
\usepackage{multirow}
\usepackage{enumitem}

\usepackage[pagebackref,breaklinks,colorlinks]{hyperref}

\usepackage[capitalize]{cleveref}
\crefname{section}{Sec.}{Secs.}
\Crefname{section}{Section}{Sections}
\Crefname{table}{Table}{Tables}
\crefname{table}{Tab.}{Tabs.}

\def\cvprPaperID{*****} 
\def\confName{CVPR}
\def\confYear{2024}

\begin{document}

\title{The Lost Melody:\\ Empirical Observations on Text-to-Video Generation\\ From A Storytelling Perspective}

\author{Andrew Shin\\
Keio University\\
{\tt\small shin@inl.ics.keio.ac.jp}
\and
Yusuke Mori\\
The University of Tokyo\\
{\tt\small mori@mi.t.u-tokyo.ac.jp}
\and
Kunitake Kaneko\\
Keio University\\
{\tt\small kaneko@dmc.keio.ac.jp}
}

\twocolumn[{%
\renewcommand\twocolumn[1][]{#1}%
\maketitle

\begin{center}
    \centering
    \captionsetup{type=figure}
    \includegraphics[width=.99\textwidth]{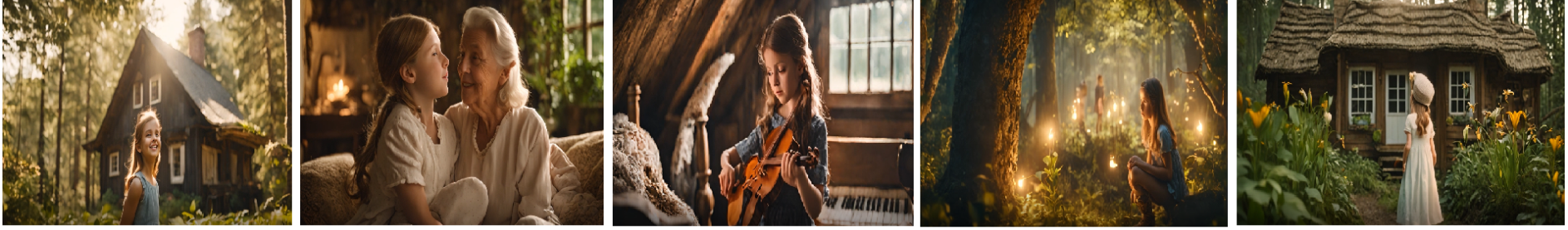}
    \captionof{figure}{Scenes from the video ``\textit{The Lost Melody}" generated from a short story.}\label{fig:main}
\end{center}%
}]

\begin{abstract}
Text-to-video generation task has witnessed a notable progress, with the generated outcomes reflecting the text prompts with high fidelity and impressive visual qualities. However, current text-to-video generation models are invariably focused on conveying the visual elements of a single scene, and have so far been indifferent to another important potential of the medium, namely a storytelling. In this paper, we examine text-to-video generation from a storytelling perspective, which has been hardly investigated, and make empirical remarks that spotlight the limitations of current text-to-video generation scheme. We also propose an evaluation framework for storytelling aspects of videos, and discuss the potential future directions.
\end{abstract}

\section{Introduction}
\label{sec:intro}

Text-to-video generation has gained attention with its increasingly better controllability, visual aesthetics, and potential for a wide range of applications, including scene generation and video editing. Creating a novel video whose duration ranges from a few seconds to a few minutes can now be done within a few minutes by simply providing text prompts, and a plethora of videos generated from text have appeared on social network services and streaming services, many of which demonstrate high-quality visuals. 

However, current text-to-video generation models almost invariably focus on generating visual components of a single scene or movement, while mostly disregarding other modalities including speech or text. As such, current text-to-video generation shares a high resemblance with movies from the silent film era, in which dialogues were either conveyed through text or completely forgone, with background music inserted to deliver the atmosphere. The difference lies in that, while most silent films strove to tell a concrete story with structured plots in spite of their lack of auditory tools, current text-to-video generation models tend to focus on presenting an array of imagery centered around the prompt, rather than storytelling. It thus becomes our central question whether contemporary text-to-video generation can also mimic the storytelling aspects of silent films that share similar limitations.  

In this paper, we examine current text-to-video generation models from a perspective of storytelling. We first examine generating videos from short stories generated by a large language model, one of which is entitled ``\textit{The Lost Melody}" (Figure~\ref{fig:main}), for which we make frequent references throughout the paper in order to illustrate important points\footnote{Sample videos are publicly available on YouTube, and can be retrieved with the title of this paper as keyword.}. We also examine generating videos from script version of each short story, in order to illustrate a striking performance drop stemming from asymmetry in training data. We further examine generating videos from captions that are aimed to perform storytelling for existing videos, in order to contrast our generated results with actual videos and captions aimed for storytelling.

Subsequently, We propose and conduct an extensive evaluation to properly assess the videos from various aspects, including components of a story, visual qualities, and how they correlate to each other. We also propose a novel evaluation framework T2Vid2T, where captions are generated from videos and are compared to input text prompts or ground truths, in a cyclical manner. Our evaluation protocols utilize both automatic and human evaluation metrics extensively, in order to account for the videos both qualitatively and quantitatively.

Upon discussion and analysis of the generation and evaluation carried out as described above, we further make empirical remarks as to the limitations that make it difficult to generate videos that successfully tell a story, and discuss potential future directions that we believe would help critically refine the performance.

Note that, since our primary focus is on storytelling, we deliberately disregard visual artifacts (which frequently appear throughout our experiments) in most cases, although our experiments suggest that they may also play a non-negligible role in storytelling. We also disregard other shooting or editing techniques frequently used in modern videos, such as changing shots and angles for more dramatic impact.

Our primary contributions can be summarized as following:
\begin{itemize}[noitemsep]
\item We examine contemporary text-to-video generation scheme from a hardly investigated storytelling perspective
\item In doing so, we propose a novel evaluation framework for videos generated from text
\item In our extensive discussion and analysis, we further suggest potential future directions for enhancing text-to-video generation in terms of reflecting storytelling aspects.
\end{itemize}

\section{Related Work}\label{sec:related_work}
\textbf{Text-to-Video Generation}:
Text-to-video has become an actively deployed research topic. Frequently employed architectures include diffusion model\cite{Ho2020DenoisingDP} and transformer\cite{Vaswani2017AttentionIA}, and also models that enable learning of common embedding for text and visual inputs, such as CLIP\cite{Radford2021LearningTV}.

Make-A-Video\cite{Singer2022MakeAVideoTG}, built on top of text-to-image generation model DALL·E 2\cite{Ramesh2022HierarchicalTI}, which in turn is based on CLIP, generates videos using spatio-temporal convolution and frame interpolation networks. Temporal corrections are made using a U-net-style diffusion model, and frame interpolation network fills in the gaps between frames generated by spatio-temporal decoder, resulting in a smoothly moving video. CogVideo\cite{Hong2022CogVideoLP}, also based on text-to-image generation model CogView2\cite{Ding2022CogView2FA}, utilizes VQVAE\cite{Oord2017NeuralDR} to convert each frame of a video into image tokens. Since applying the same frame rate to all videos can lead to mismatches between the content of the videos and the text, and consecutive frames tend to have very similar content, making it challenging to learn long-term dependencies with a fixed frame rate, CogVideo inserts a token indicating the frame rate into the text and samples frames at the specified frame rate. Imagen Video\cite{Ho2022ImagenVH} utilizes a cascade of spatio-temporal super-resolution, generating a total of 128 frames with a resolution of 1280×768 at a frame rate of 24fps. The advantage of such cascade model is that it can independently learn each super-resolution diffusion model. The text encoder uses a pre-trained T5\cite{Raffel2019ExploringTL} model with fixed weights. Imagen Video is particularly noteworthy for its ability to represent text within videos, a task that was challenging for traditional video generation models.

Sora\cite{videoworldsimulators2024} is a diffusion model using transformer architecture, and enables generation of videos up to 1 minute that demonstrate high fidelity to input prompts with highly plausible graphics. Recaptioning technique from DALL·E 3\cite{OpenAI2023} has been said to have helped generating videos with high fidelity to users' text input prompts.

While many text-to-video generation models do not allow for public examination and explicitly state the limitations, it is generally fair to say that most of the text-to-video generation models above presuppose that the input textual prompt aims to describe a single scene or motion, such that, upon receiving prompts that contain multiple scenes as necessary for a storytelling, the models usually end up reflecting only a limited portion of the input, failing to generate results that successfully perform storytelling.
Its causes may be attributed to various factors including limitations on input length, duration of output, and training data.

\textbf{Visual Storytelling}: 
As a more advanced task than conventional image captioning task, visual storytelling \cite{Huang2016VisualS} was proposed to generate a story based on a sequence of images. In contrast, \cite{li2018storygan} proposed the story visualization task, attempting to generate a sequence of images from a story. Examining video generation from stories is a natural extension of the line of works above that attempt to bridge the gap between vision and storytelling. Recent works have tackled the task of consistent multi-scene text-to-video generation\cite{Lin2023VideoDirectorGPTCM, Long2024VideoDrafterCM}, but their focus is on multi-scene generation from a single prompt, rather than a wider scope of storytelling.


\textbf{Story Evaluation}:
Automatic evaluation of stories is an essential research topic for tasks related to stories.
However, in textual story evaluation, it has been pointed out that existing metrics correlate poorly with human evaluation~\cite{guan-etal-2021-openmeva}. In order to overcome such low correlation, novel evaluation metrics including UNION~\cite{guan-huang-2020-union} and StoryER~\cite{chen-etal-2022-storyer} have been proposed and demonstrated improved correlations with human perception. 

Such limitation of conventional evaluation metrics has also been pointed out in visual storytelling\cite{hsu-etal-2019-visual,wang-etal-2018-metrics}.
\cite{Hu2019WhatMA} proposed a metric for visual storytelling from three perspectives; namely \textit{relevance}, \textit{coherence}, and \textit{expressiveness}.
Inspired by \cite{Hu2019WhatMA}, \cite{wang-etal-2022-rovist} proposed the similar three perspectives to evaluate visual storytelling; \textit{visual grounding}, \textit{coherence}, and \textit{non-redundancy}. To assess story quality, \cite{liu-etal-2023-visual-storytelling} used story-specific metrics in addition to lexical-matching metrics.


\section{Components of a Story}\label{sec:components}

While various definitions exist as to what constitutes a story, common elements include character, plot, setting, conflict, theme, and point of view. We briefly describe each component and discuss the challenges it poses on videos generated from text, which will also be examined explicitly later in the paper.

\textbf{Character} is the person in the story, around which the story evolves. In this simple sense, we can say that videos generated from text also frequently contain characters. However, there is rarely a single character in a story, and multiple characters including a protagonist, the main character, and an antagonist, an opposition to the protagonist, are typically present. Furthermore, character can also refer to the qualities of a person. Direct characterization, where the qualities are directly presented to the audience, such as their appearances, can be easily manipulated in videos generated from text, and certain aspects of their personalities can often be specified by adjusting facial expressions. Indirect characterization, however, where the qualities are formed through the person's statements or behaviors, inevitably necessitates a development of a plot. 

\textbf{Setting} refers to time and location in which the story takes place. The degree of specification required varies depending on the story, and videos generated from text inherently provide direct visual clues for the setting. Also, setting may contain the social conditions under which the character is placed, which is presented less directly and requires context. Although not entirely impossible, it is certainly more challenging for videos generated from text to accurately present.

\textbf{Plot} refers to the sequence of events that occur throughout the story, and develops through multiple stages including introduction, rising action, climax, falling action, and resolution. While it is certainly possible in theory to demonstrate these multiple stages solely with visual elements, as was the case with silent movies, it turns out to be difficult with current text-to-video generation, as we will see later. 

Conflict, often considered to be an element of plot, refers to the primary opposition that the character undergoes, and may be internal, \textit{i.e.} versus self, or external, \textit{e.g.} with other characters or circumstances. In any case, insomuch as the plot is nearly absent, it is evident that conflict is also difficult to convey. Theme refers to the central message of the story, and may not always be directly expressed. While visual elements often strongly contribute to the highlighting of a theme, it is typically developed through the plot, e.g. the outcomes of a character's behaviour. As such, it is highly challenging for current text-to-video generation to effectively express a theme. Point of view in a story of text format refers to the narrator; first person if the narrator is a character within the story, and third person if the narrator is not. Point of view takes a slightly different meaning when applied to videos. Implementing a truly first person's point of view would be to use views seen from the character's eyes, and while such shot is rare in commercial films and television shows, it can be generated by specifying it in input prompts. Third person's point of view, where the audience oversees what is happening as an observer, is in most cases the default point of view for videos generated from text, unless specified otherwise.

In addition, other elements may be included, such as style or tone of the story, and symbolism. The former can be conveyed through the visual atmosphere of the video, which can also be controlled with specifications in input prompts. The latter is more of a derivative of a plot, and thus can hardly exist in the absence of a plot.

In summary, current text-to-video generation, while inherently containing some elements of a story to varying extents, may struggle to fully attain the indispensable elements that constitute a story, as we will empirically observe later. In this paper, we primarily focus on character, setting, and plot, as other components can be considered sub-components of one of these, and without accounting for the those three components, reflecting the remaining components will be simply out of question.

\section{Video Generation}

\begin{figure*}[ht]
  \centering
  \includegraphics[width=0.8\linewidth]{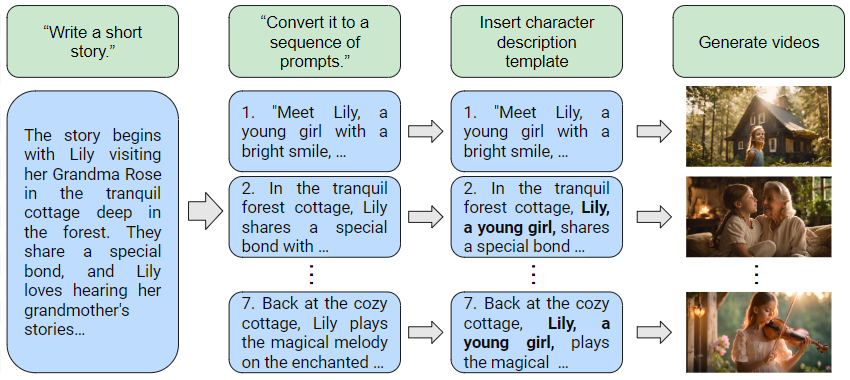}
  \caption{Overall workflow of generating videos from a short story generated by a large language model.}
  \label{fig:pipeline}
  \vspace{-5mm}
\end{figure*}
We now describe our workflow of generating videos for stories. We examine generation from three distinct types of text prompts; short stories generated by LLM, scripts generated by LLM, and captions that describe existing videos.

\subsection{Generation from Short Story}
We used ChatGPT\cite{OpenAI2022} with GPT3.5 to generate a short story from which to generate a video. The generated short story consists of multiple scenes. As discussed in Section~\ref{sec:related_work}, with current text-to-video generation, it is nearly infeasible to generate a video for the entire story from a single round of generation, in terms of both maximum token length acceptable and the model's ability to compose multiple scenes. As such, we need to convert the generated short story into a sequence of prompts, where each prompt corresponds to a scene. Concatenating the generated video for each prompt will be the resulting video of the short story. 

Note that each scene is generated independently of each other, without any explicit adherence to previously generated outcomes. As such, each prompt for video generation model needs to specify the characters and the setting repeatedly. Otherwise, the context is easily lost and the generation results in incoherent videos. An alternative would be to generate the videos conditioned on previously generated outcomes. We did examine conditioning video generation on an image from previously generated results, but the result turned out to be unreliable as we will further discuss in Section~\ref{sec:discussion}. 

\begin{figure}[t]
  \centering
  \includegraphics[width=0.9\linewidth]{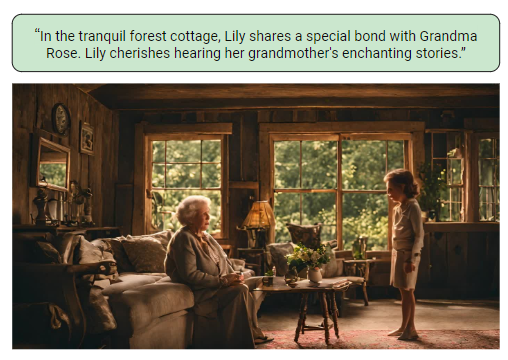}
  \caption{Unless each prompt contains a sufficient amount of details, the model generates incoherent results.}
  \label{fig:insufficient_prompt}
  \vspace{-5mm}
\end{figure}

Since each scene is generated independently, we also insert character description for proper nouns repetitively in each prompt. For example, "\textit{Lily shares a special bond with ...}" is replaced with "\textit{Lily, a young girl, shares a special bond with...}," and so forth for each prompt. Otherwise, apart from the first appearance of the character, none of the visual descriptions will be available in subsequent scenes, resulting in immense inconsistency, as illustrated in Figure~\ref{fig:insufficient_prompt}. As we will re-visit in Section~\ref{sec:discussion}, a scheme to integrate previous context, frequently used in LLMs, may help achieve visual coherence more easily.

Finally, video is generated for each prompt with Gen-2\cite{Esser2023StructureAC}, after which the generated videos are temporally concatenated. In order to examine how the comprehensibility of the story changes with linguistic aid, we generate speech from each prompt using OpenVoice\cite{Qin2023OpenVoiceVI}. We adjuste the playing speed of each scene in order to match the duration of narration. The same duration is applied to the version without narration, as the varying duration can affect the viewer's understanding of the story. For aesthetic purpose, we also generate the background music with Stable Audio\cite{Evans2024FastTL} using the scene description as the prompt, which was applied to all of our videos in an identical manner. 

Figure~\ref{fig:pipeline} shows the overall workflow of generating videos for a short story created by an LLM. See Supplemental Material for actual prompts we used and the generated outcomes. 

\subsection{Generation from Script}\label{subsec:gen_script}
Text-to-image generation or text-to-video generation have mostly evolved around descriptive text prompts, most of which are concerned with vividly conveying the visual components to be generated. Yet, humans can also visualize the scene from dialogue, which is in fact a strong medium for conveying a narrative along with novel. In fact, dramas or plays written in script format have longer history than novels, and have been around for millennia. In order to examine to what extent video generation can handle storytelling given text prompt in script format, we generate a script version of the short story generated as described above. We use LLM to generate a sequence of prompts for video generation model, explicitly requiring the dialogues spoken by characters, with non-dialogue parts for indicating setting or transitions. As before, each scene is generated independently, so proper nouns in each dialogue are supplemented with description for each prompt.

\subsection{Generation from Captions}
In order to make a comparison with how story is perceived from real videos, we sampled a subset from Video Storytelling dataset\cite{Li2018VideoST}, whose ground truth captions were used for video generation. While other datasets that examined visual storytelling exist, such as SIND\cite{Huang2016VisualS}, most of them contain a sequence of images rather than videos, and are thus not apt for our purpose. Since videos in Video Storytelling dataset usually span several minutes, which are lengthier than our generated videos, we either used the first 5 captions, or the first 100 seconds of the video, selecting whichever one results in shorter duration. 

\section{Experiments}
\subsection{Evaluation}
As shown in Section~\ref{sec:related_work}, while there have been evaluation metrics proposed for textual storytelling or video captioning, evaluation of videos from a storytelling perspective has hardly been examined, to the best of our knowledge. In this section, we propose and conduct a wide range of evaluation metrics to assess storytelling ability of videos both qualitatively and quantitatively, involving both automatic and human evaluations.

We first evaluate the visual quality of the generated videos with FVD (Fréchet video distance)\cite{Unterthiner2018TowardsAG} and inception score (IS)\cite{Salimans2016ImprovedTF}. For FVD, following \cite{Clark2019AdversarialVG}, we used I3D\cite{Carreira2017QuoVA} trained on Kinetics-600\cite{Carreira2018ASN} to compute the real video statistics. For inception score, following \cite{Saito2016TemporalGA}, we used pre-trained C3D\cite{Tran2014LearningSF}, first trained on Sports-1M\cite{Karpathy2014LargeScaleVC} and fine-tuned on UCF101\cite{Soomro2012UCF101AD}. We extracted 10 frames per second from the target videos and report the average values. While these metrics are not directly relevant to storytelling \textit{per se}, we will later use them to examine how they correlate to the audience's perception of the story.

\begin{table}[t]
\caption{FVD and inception score on each type of videos. Lower FVD and higher inception score indicate better results.}
    \label{tab:fvd_is}
    \centering
    \begin{tabular}{c|c|c}
        \hline
        Model & FVD($\downarrow$) & Inception Score($\uparrow$) \\
        \hline
        Generated (story)& 51.53 & 11.01 \\
        Generated (script) & 75.34 & 10.25 \\ \hline
        Generated (caption) & 46.06 & 12.22 \\
        Real videos & \bf 6.54 & \bf 32.18 \\
        \hline
    \end{tabular}
    \vspace{-4mm}
\end{table}

\begin{figure*}[ht]
  \centering
  \includegraphics[width=0.8\linewidth]{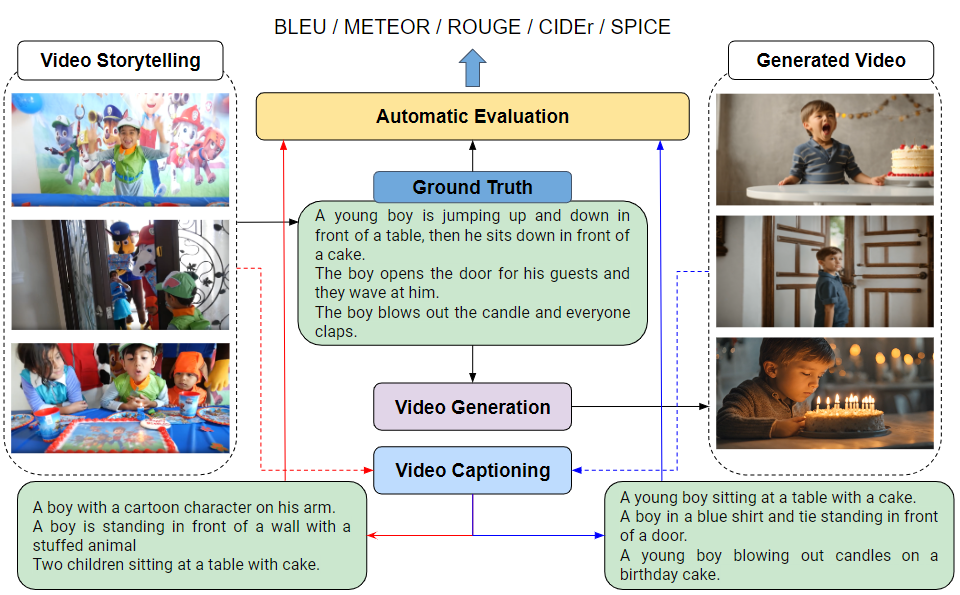}
  \caption{Evaluation workflow for T2Vid2T.}
  \label{fig:t2vid2t}
  \vspace{-4mm}
\end{figure*}

We also propose a novel cyclical evaluation framework, namely T2Vid2T (text-to-video-to-text), where we automatically generate captions that describe the generated video, and compare it against the original text prompt that was used to generate the video. We used a fine-tuned version of TimeSFormer\cite{Bertasius2021IsSA} for video captioning to generate captions from each scene. Popular evaluation metrics including BLEU\cite{Papineni2002BleuAM}, METEOR\cite{Denkowski2014MeteorUL}, ROUGE\cite{Lin2004ROUGEAP}, CIDEr\cite{Vedantam2014CIDErCI}, and SPICE\cite{Anderson2016SPICESP} were employed for comparison against text prompt. Note that we replace the proper nouns in text prompt with general descriptive phrase, e.g. "\textit{Lily}" with "\textit{a young girl}", as the proper noun is out of scope with video captioning framework. For Video Storytelling, we disregarded the parts where ground truth captions were not provided, and extracted only the frames that were accompanied by captions. Figure~\ref{fig:t2vid2t} illustrates the workflow of T2Vid2T using Video Storytelling as the reference. 

\begin{table*}[t]
\caption{Results of T2Vid2T}
    \label{tab:t2vid2t}
    \centering
    \begin{tabular}{c|c|c|c|c|c}
        \hline
        Model& BLEU-1 & METEOR & ROUGE-L & CIDEr-D & SPICE \\
        \hline
        Generated (story)& 0.0298 & 0.0343 & 0.1525 & 0.0275 & 0.0877\\
        Generated (script) & 0.0298 & 0.0258& 0.1325 & 0.0011& 0.0440\\ \hline
        Generated (caption) & 0.1045 & 0.0635 & 0.1244 & 0.0615& 0.0952\\
        Real videos & \bf 0.2630 & \bf 0.1683 & \bf 0.2572 & \bf 0.2604 & \bf 0.3519\\
        \hline
    \end{tabular}
    \vspace{-4mm}
\end{table*}

We further conducted human evaluation, where we asked the workers to rate the video in terms of the components of the story, namely character, setting, plot, as outlined in Section~\ref{sec:components}, out of 1 to 5 scale. Each component of the story was rated according to three distinct aspects, namely \textit{expressiveness}, \textit{relevance}, and \textit{coherence}, as suggested by \cite{Hu2019WhatMA}, yielding a $3\times 3$ evaluation matrix per video. \textit{Expressiveness} refers to richness and diversity of expression styles. \textit{Relevance} refers to the extent to which the video reflects the text prompt in terms of the story component of concern. \textit{Coherence} refers to the extent to which the consistency of respective component is retained throughout the video. In addition, a worker was asked to rate the video in terms of overall comprehensibility of the story.

\begin{table*}[t]
\caption{Ratings for generated videos in terms of the components of a story. Numbers in the parenthesis indicate the ratings when the workers were asked to take visual artifacts into consideration as well.}
    \label{tab:human}
    \centering
    \scalebox{0.82}{
    \begin{tabular}{c|*{9}{c|}c}
        \hline
        \multicolumn{1}{c|}{\multirow{2}{*}{Model}} & \multicolumn{3}{c|}{Character} & \multicolumn{3}{c|}{Setting} & \multicolumn{3}{c|}{Plot} &Overall\\
        \cline{2-10}
        & Exp. & Rel. & Coh. & Exp. & Rel. & Coh. & Exp. & Rel. & Coh.& Comprehensibility \\
        \hline
        Gen. (story) & 2.5(2.2) & 2.0(1.7) & 2.8(2.6) & 3.0(2.5) & 3.5(3.2) & 2.8(2.6) & 3.2(2.8) & 3.2(3.1) & 2.9(2.4) &2.7(2.0)\\
        Gen. (story) (narration) & 2.7(2.2) & 1.8(1.7) & 2.9(2.8) & 3.3(2.9) & 3.7(3.2) & 3.0(2.5)& 3.4(3.0) & 3.0(3.0) & 3.0(2.6) &3.2(2.5)\\
        Gen. (script) & 2.8(2.5) &1.7(1.6) & 2.1(2.0) &2.8(2.6) &2.5(2.2) &1.5(1.3) & 1.9(1.8) & 2.0(1.9) & 2.2(1.8)& 2.1(1.9)\\
        Gen. (script) (narration) & 2.8(2.6) & 1.4(1.2) & 2.0(1.5) & 2.8(2.4) & 2.6(2.4) & 1.7(1.6) & 2.0(1.8) & 2.2(2.2) & 2.5(2.2)& 2.5(2.2)\\\hline
        Gen. (caption) & 3.5(3.2) & 3.8(3.4) & 3.6(3.3)& 2.5(2.4) & 3.2(2.7) & 3.0(2.7) & 3.5(3.0) & 3.9(3.6) & 3.7(3.3) &3.8(3.6)\\
        Gen. (caption) (narration)& 3.7(3.7) & 4.0(3.8) & 3.8(3.7) &2.8(2.4) & 3.2(3.0) & 3.1(2.7) & 3.7(3.7) &4.2(3.9) & 3.7(3.7) &4.0(4.0)\\
        Real videos & \bf 4.6 &\bf 4.7 & \bf 4.8 & \bf 4.5 & \bf 4.2 & \bf 4.0 & \bf 3.8 & \bf 4.8 & \bf 4.9 &\bf 4.8\\
        \hline
    \end{tabular}
    }
    \vspace{-4mm}
\end{table*}

We used Amazon Mechanical Turk for our experiment. We presented the workers with videos generated from a short story, a script version of the story, and a caption from Video Storytelling, along with an actual video to which the caption belongs. Each generated video is presented in two variations; the generated video as is, and the generated video with narration added to it. Note that human evaluation was performed with 2 conditions, where the worker was either explicitly asked to ignore the visual artifacts, or to take visual artifacts into consideration, in order to examine how they contribute to perception of visual storytelling.

Finally, we conducted a mixture of human and automatic evaluation, where the worker was asked to write a summary of the video, with one sentence per scene, and the resulting summary was compared against the input textual prompt using automatic evaluation metrics.

\begin{table*}[t]
\caption{Results of automatic evaluations on human-written summary.}
    \label{tab:human_summary}
    \centering
    \begin{tabular}{c|c|c|c|c|c}
        \hline
        Model & BLEU-1 & METEOR & ROUGE-L & CIDEr-D & SPICE \\
        \hline
        Generated (story)& 0.0444 & 0.0595 & 0.1614 & 0.0038 &0.1607\\
        Generated (script) & 0.0348 & 0.0537& 0.1570 & 0.0036&0.1326\\ \hline
        Generated (caption) & 0.2467 & 0.1499 & 0.3377 & 0.7429 & 0.2889\\
        Real videos & \bf 0.3096 & \bf 0.2299 & \bf 0.5137 & \bf 1.6115 &  \bf 0.3060\\
        \hline
    \end{tabular}
    \vspace{-4mm}
\end{table*}

\begin{table*}[t]
\caption{Examples of human-written summary. The first two summaries are based on ``\textit{The Lost Melody}", and the last two are based on the birthday episode shown in Figure~\ref{fig:t2vid2t}.}
    \label{tab:summary_example}
    \centering
    \begin{tabular}{c|l}
        \hline
        Model & Human-written summary \\
        \hline
        {\multirow{4}{*}{Gen.(story)}}&``A little girl is smiling in front of a house in forest. The girl is sitting next to her grandma. The girl is\\& playing a violin in a room. Now, the girl goes outside and starts playing violin again. The girl is lost in\\& the forest and people are searching for her. The girl manages to come back home safely. The girl plays\\& violin again with joy of returning home."\\
 \hline
        {\multirow{3}{*}{Gen.(script)}} &``A young girl and grandma are enjoying their time next to fireplace. The girl starts playing a violin. The\\& girl goes out to forest. In the forest, she is fascinated by nature around her. Back at home, the girl and\\& her mother start growing flowers. Now she plays violin with a different meaning of loving nature."  \\ \hline
        {\multirow{2}{*}{Gen.(caption)}}  & ``A boy is happy with his birthday cake. The boy opens the door. The boy, with his eyes closed, is\\& praying in front of the cake with candles lit."\\\hline
        {\multirow{2}{*}{Real}} & ``A happy boy is jumping in front of his birthday cake.
The boy opens the door to welcome his friends.\\&
The boy blows the candles and the friends celebrate." \\
        \hline
    \end{tabular}
    \vspace{-4mm}
\end{table*}



\subsection{Results}\label{subsec:results}
Table~\ref{tab:fvd_is} shows FVD and inception score for each type of videos. Real videos from Video Storytelling outperform other generated videos by far. While such result is highly predictable, it is particularly of our interest that higher visual qualities also correlate to perception of story, as will be shown in other results. Also, videos generated from captions for Video Storytelling displayed better results than videos generated from short stories or scripts, suggesting that current text-to-video generation models are more compatible with factual descriptions.

Table~\ref{tab:t2vid2t} shows the results of T2Vid2T. Note that, input prompts were used as ground truths for results other than Video Storytelling.
 Video captioning is still widely considered a relatively difficult task, as can also be seen by notably low scores throughout all videos. Setting aside the real videos, which performed far better as predicted, it is notable that videos generated from stories and scripts performed substantially poorly, even when compared to videos generated from captions. This suggests that video captioning model struggles even to match a few tokens with the input stories or scripts, hinting at a severe imbalance in the training corpora in terms of textual styles and formats.

\begin{figure}[t]
  \centering
  \includegraphics[width=0.9\linewidth]{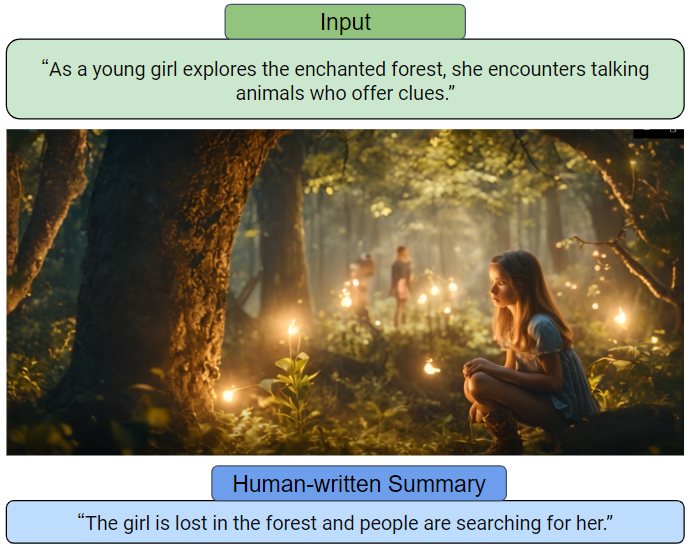}
  \caption{Example of human-written summary that is substantially different from input. }
  \label{fig:misunderstanding}
  \vspace{-4mm}
\end{figure}

Table~\ref{tab:human} summarizes the results from human evaluation in terms of how well the video reflects components of a story. While trends from previous results generally continue, adding linguistic elements via narration improved the overall scores. It is notable that in some categories, videos with narrations received lower scores than videos without narrations. We conjecture that this is due to mismatch between narration and the generated scene. For example, text-to-video generation frequently struggles with multiple characters, and easily ends up generating incoherent characters or skipping the generation of some characters present in input prompt. Such mismatch is likely to exacerbate, rather than aid, the perception of story. Asking the workers to take visual artifacts into consideration resulted in performance drop, suggesting that visual qualities also play a role in perception of stories. In fact, our investigation showed that visual artifacts also had an influence on automatic evaluations by incorrectly classifying the objects, \textit{e.g.} violin to guitar, etc. Table~\ref{tab:human_summary} shows the results of running automatic evaluation metrics on human-written summary upon watching the videos, with input prompts as the ground truths. Its overall trend is coherent with Table~\ref{tab:t2vid2t}, suggesting that the low performance of videos generated from stories and scripts is not solely attributed to problems with video captioning models, but is also highly relevant to limitations with text-to-video generation models as well. Table~\ref{tab:summary_example} shows example human-written summary for each type of videos. Note that, for ``\textit{The Lost Melody}", later parts of the plot were often completely misunderstood by the workers. This reinforces our claims and experiment results that current text-to-video generation models are yet to generate convincing videos for storytelling. Figure~\ref{fig:misunderstanding} visualizes how the plot for a scene was misunderstood by the worker.

\section{Discussion}\label{sec:discussion}
In this section, we review some of the notable drawbacks of current text-to-video generation models in terms of storytelling as observed in our experiments, and discuss potential future directions that would enhance the performance.

\begin{figure}[t]
  \centering
  \includegraphics[width=0.9\linewidth]{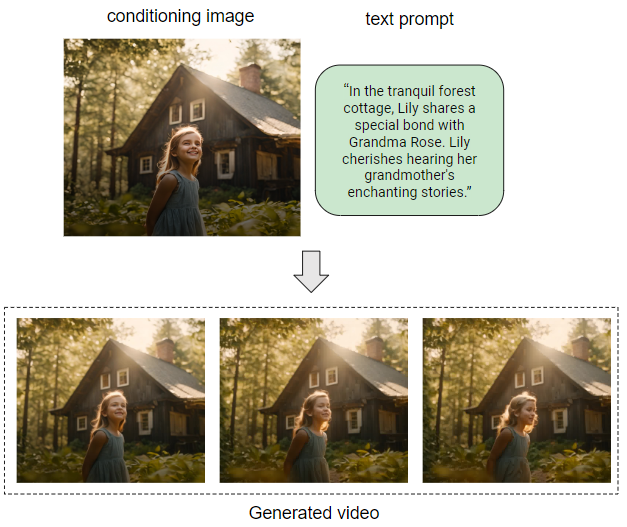}
  \caption{A failure case of video generation conditioned on both image and text inputs.}
  \label{fig:conditioning_fail}
  \vspace{-4mm}
\end{figure}

\textbf{Video generation with visual reference:}
We generated videos for the story simply by concatenating multiple generated videos for its constituents. Such workflow is bound to incur inconsistency in visual aspects, as we have witnessed in our experiments. An alternative would be to condition the text-to-video generation on a reference visual, from which to learn how the characters should look like, and reflect it in the generated videos. Indeed, a story generation scheme with additional source frame has been proposed for text-to-image generation\cite{Maharana2022StoryDALLEAP}, and some of the text-to-video generation models also enable both image and text inputs. However, it must be noted that, in storytelling perspective, the purpose of conditioning the generation on image and text is fairly different from current editing schemes based on image and text inputs. In text-based image editing, for example, input images themselves are targets to be directly modified via directions in the text prompt. A direct extension of this to video generation, which is currently deployed by some of the text-to-video generation models, also entails an assumption that input images are targets to be directly modified in the generated video, \textit{
e.g.} generating motions on the image as specified by input prompt. As such, when confronted with text inputs that do not imply a direct editing of the input image, the resultant video utterly fails to reflect any component of the input prompt, and merely ends up making subtle motions on the input image (see Figure~\ref{fig:conditioning_fail}).
In storytelling perspective, image or video inputs conditioned for the generation are frequently not the targets to be directly modified, but are references that provide clues as to how the main components of the video should appear, so that the consistency of visual storytelling is preserved. As such, a scheme to condition the video generation on image not by directly modifying it, but using it as a reference to adjust the visual appearances will be necessary to enable a recurrent video generation, which is more desirable than current scheme of concatenating independently generated scenes.


\begin{figure}[t]
  \centering
  \includegraphics[width=0.9\linewidth]{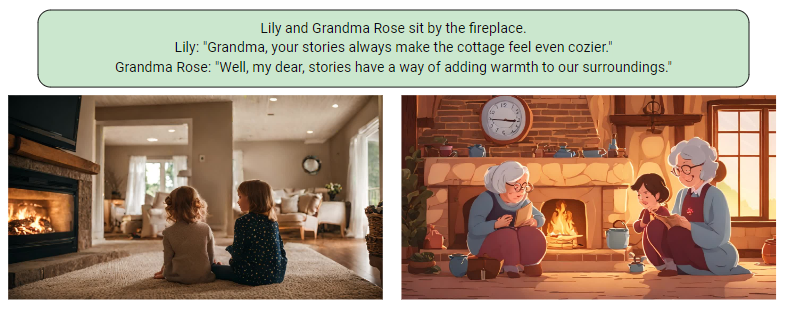}
  \caption{Generating videos from script format frequently results in complete failure of understanding the scene.}
  \label{fig:dialogue_failure}
  \vspace{-4mm}
\end{figure}

\textbf{Script-to-video generation:} 
As discussed in Section~\ref{subsec:gen_script}, script is a familiar format for humans to perceive stories. As shown in Section~\ref{subsec:results}, nonetheless, generating videos from dialogue is as of now far from delivering a sensible and coherent visuals, which implies a severe asymmetry in terms of the types of text in training corpus. In our attempts, the model frequently fails with who is present in the scene, when given a script as input prompt, as shown in Figure~\ref{fig:dialogue_failure}. This implies that the current model does not understand how the script format is structured, particularly the role of speaker identification token. Once dialogue-to-video generation is established, a much more powerful storytelling will be possible, as it can be more easily incorporated with audible modality using text-to-speech generation, which turned out to be helpful in conveying a narrative in our experiments. Lip-syncing the characters to given text is also an active research topic\cite{Kumar2017ObamaNetPL}, and can be more easily integrated with videos if dialogue is given.

\textbf{Global story representation: }
Our premise so far has been a concatenation of separately generated videos. Ideally, however, a scheme for a global understanding of structure of the story will be necessary. In music generation, for example, a hierarchical architecture is frequently employed\cite{Roberts2018AHL} in order to learn the global structure of a musical piece, consisting of ascendance, descent, and repetition of theme, which are also observed in stories. Näively porting the same approach to video domain, however, will require prohibitively large amount of computational resources, and it may thus be necessary to come up with a resource-efficient approach that can compactly encode the structure of a story. One possibility is to represent the structure of a story with graphs, as have been examined by previous works\cite{Xu2024GraphRO,Lukin2016PersonaBankAC}. Incorporating such global representation into video generation can potentially enhance the generation of more coherent videos.

\textbf{Evaluation metrics}: While we proposed and conducted a number of evaluation protocols in this paper, they still relied heavily on manual efforts by human. Fully automatic evaluation metrics, such as T2Vid2T, also suffered from still very low similarity between generated captions and ground truth stories, and it also does not account for visual consistency of the character. Such issues may be mitigated by building a more story-oriented video captioning model, which in turn requires a construction of large-scale datasets with videos and corresponding stories. In any case, it is inarguable that more reliable and standardized evaluation metrics for storytelling aspects of videos are indispensable for the field to make substantial advances.

\section{Conclusion}
We examined current text-to-video generations from a novel storytelling perspective, by generating videos from short stories and scripts, as well as existing captions for video storytelling. We performed an extensive evaluation of generated videos by proposing and conducting a wide array of evaluation protocols. Finally, we made a number of empirical observations and suggestions for potential future directions that we believe would help enhance text-to-video generation for storytelling.

{\small
\bibliographystyle{ieee_fullname}
\bibliography{egbib,tmp_mori_add}
}

\appendix
\section{Appendix}

\subsection{Short Story Generation}
\textbf{Prompt:} 

Write a plot for a 5-minute animation. 
It should involve basic elements of a story, including characters, setting, plot, conflict, and theme.
The overall mood should be warm and heartfelt, but it may be good to add an element of eeriness.

\textbf{Output:}

Title: "The Lost Melody"

Setting: A cozy, rustic cottage in a peaceful forest.

Characters:

Lily: A young girl with a bright smile, staying at the cottage with her grandmother.
Grandma Rose: A kind and wise elderly woman who lives in the cottage.
The Enchanted Violin: An old, mysterious violin found in the attic.
Plot:

The story begins with Lily visiting her Grandma Rose in the tranquil cottage deep in the forest. They share a special bond, and Lily loves hearing her grandmother's stories.

One day, while exploring the attic, Lily stumbles upon an old, dusty violin. Grandma Rose explains that it's an enchanted violin, but it has been silent for years because it's missing a magical melody.

Intrigued by the idea of an enchanted violin, Lily is determined to find the lost melody. She embarks on a quest through the forest, seeking magical elements to create the melody. Along the way, she encounters talking animals who offer clues, and the forest itself seems to come to life.

As Lily's journey unfolds, an eerie, mystical atmosphere envelopes the forest. Strange and enchanting occurrences, like glowing fireflies and ethereal music, add an element of eeriness to the story.

Finally, with the help of the forest creatures, Lily finds all the ingredients needed for the magical melody and returns to the cottage.

She plays the melody on the enchanted violin. It responds with beautiful, heartwarming music that fills the forest, causing flowers to bloom and animals to dance.

The message of the story becomes clear: The magic is in the connection between generations and the beauty of passing down traditions and stories.

Theme:
The central theme of "The Lost Melody" is the power of tradition, the importance of family bonds, and the enchanting wonder that can be found in even the most unexpected places. It combines warmth and heartfelt moments with an element of mystery and magic, creating an unforgettable and heartwarming tale.

\subsection{Prompt Sequence Generation}
\textbf{Prompt:}

I want to generate a video for the generated short story above using AI text-to-video generation tools. So I need to divide them into a sequence of texts, each of which is to be the prompt for the generation. For the prompt to successfully generate coherent video, each prompt will have to contain the character and setting, since each prompt is independent of others from the tool's perspective. Please make the prompt sequences.

\textbf{Output:}

Introduction to Setting and Character:

"Meet Lily, a young girl with a bright smile, staying at a cozy, rustic cottage nestled in a peaceful forest. Lily is visiting her grandmother, Grandma Rose."

Special Bond and Grandma's Stories:
"In the tranquil forest cottage, Lily shares a special bond with Grandma Rose. Lily cherishes hearing her grandmother's enchanting stories."

Discovery of the Enchanted Violin:
"While exploring the attic one day, Lily discovers an old, dusty violin in the cozy cottage. Grandma Rose reveals its mystery – it's an enchanted violin, silent for years without its magical melody."

Quest for the Lost Melody Begins:
"Intrigued by the enchanted violin, Lily is determined to find the lost melody. She embarks on a quest through the mystical forest surrounding the cozy cottage, seeking magical elements to create the melody."

Encounters in the Enchanted Forest:
"As Lily explores the enchanted forest, she encounters talking animals who offer clues. The forest itself seems to come to life around the cozy cottage. Strange occurrences, like glowing fireflies and ethereal music, add an eerie yet enchanting atmosphere."

Culmination of Lily's Journey:
"During Lily's journey, an eerie, mystical atmosphere envelops the forest around the cozy cottage. With the help of the forest creatures, Lily gathers all the magical ingredients and returns to the cottage."

Magical Melody Unveiled:
"Back at the cozy cottage, Lily plays the magical melody on the enchanted violin. The violin responds with beautiful, heartwarming music that fills the forest. Flowers bloom, and animals dance in response."

The Message of the Story:
"In this enchanting tale, discover the magic in the connection between generations and the beauty of passing down traditions and stories at the cozy, rustic cottage in the peaceful forest."
Now, each prompt contains both the character (Lily, Grandma Rose) and the setting (cozy cottage in the peaceful forest). Adjustments can still be made based on the specific requirements of the text-to-video generation tool.

\subsection{Script Generation}
\textbf{Prompt:}

Convert the story above into a script with dialogue to be spoken by characters, with non-dialogue part for indicating setting, transition, etc. Write the script as a prompt sequence for AI text-to-video generation tools, while including characters in each prompt as you did before.

\textbf{Output:}

INT. COZY COTTAGE - DAY

Setting: A cozy, rustic cottage in a peaceful forest.
Characters: Lily (young girl), Grandma Rose (elderly woman)
INT. COZY COTTAGE - LIVING ROOM - DAY

Lily and Grandma Rose sit by the fireplace.
Dialogue:
Lily: "Grandma, your stories always make the cottage feel even cozier."
Grandma Rose: "Well, my dear, stories have a way of adding warmth to our surroundings."
INT. COZY COTTAGE - ATTIC - DAY

Lily explores the attic, discovering an old, dusty violin.
Dialogue:
Lily: "What's this, Grandma?"
Grandma Rose: "Ah, that, Lily, is no ordinary violin. It's enchanted, but it has been silent for years."
EXT. COZY COTTAGE - DAY

Lily sets out on a quest through the mystical forest.
INT/EXT. ENCHANTED FOREST - VARIOUS LOCATIONS - DAY

Lily encounters talking animals, gathers magical elements, and experiences the mystical atmosphere of the forest.
Dialogue:
Lily (to a talking bird): "Have you heard of the lost melody? I need your help."
Talking Bird (chirping): "Follow the fireflies. They hold the secret."
EXT. COZY COTTAGE - DAY

Lily returns to the cottage, the enchanted forest echoing with mysterious sounds.
INT. COZY COTTAGE - LIVING ROOM - DAY

Lily, with gathered ingredients, stands before Grandma Rose.
Dialogue:
Lily: "Grandma, I'm going to play the lost melody on the enchanted violin!"
Grandma Rose: "Oh, Lily, that melody has been silent for far too long. Let the forest hear its song again."
EXT. COZY COTTAGE - DAY

Lily plays the magical melody. The forest responds with blooming flowers and dancing animals.
INT. COZY COTTAGE - LIVING ROOM - DAY

Grandma Rose watches with joy.
Dialogue:
Grandma Rose (whispering): "The magic is in the connection between generations, Lily. You've brought the melody back to life."
EXT. COZY COTTAGE - DAY

The forest, now alive with enchantment, surrounds the cozy cottage.

\end{document}